\documentclass[preprint,review,5pt]{elsarticle}
\usepackage{graphicx}
\usepackage{amsmath,amssymb}
\usepackage{lineno}
\usepackage{color}
\usepackage{subfigure}
\usepackage{algorithm}
\usepackage{txfonts}
\usepackage{multirow}

\begin{document}
\begin{frontmatter}
\title{An Aerial Image Recognition Framework using Discrimination and Redundancy Quality Measure}
\author{Yuxing Hu and Luming Zhang \\
School of Computing, National University of Singapore}

\begin{abstract}
Aerial image {categorization plays an indispensable role in remote
sensing and artificial intelligence.} In this paper, we propose a
new aerial image categorization framework, focusing on organizing
the local patches of each aerial image into multiple
discriminative subgraphs. {The subgraphs reflect both the
geometric property and the color distribution of an aerial image.}
First, each aerial image is decomposed into a collection of
regions in terms of their color intensities. Thereby region
connected graph (RCG), which models the connection between the
spatial neighboring regions, is constructed to encode the spatial
context of an aerial image. Second, {a subgraph mining technique
is adopted to} discover the frequent structures in the RCGs
constructed from the training aerial images. {Thereafter, a set of
refined structures are selected among the frequent ones toward
being highly discriminative and low redundant.} Lastly, given a
new aerial image, its sub-RCGs corresponding to the refined
structures are {extracted. They are further quantized into a
discriminative vector for SVM classification.} Thorough
experimental results validate the effectiveness of the proposed
method. {In addition, the visualized mined subgraphs show that the
discriminative topologies of each aerial image are discovered.}
\end{abstract}

\end{frontmatter}

\section{Introduction}
Aerial image categorization is an {important
component for many applications in artificial intelligence and
remote sensing~\cite{add1,add2,add3},} such as visual
surveillance, navigation, and robot path planning. However, it is
still a challenging task to deal with aerial image categorization
successfully due to two reasons. On one hand, the aerial image
components {(\textit{e.g.}, house roofs and
grounds) as well as their spatial configurations are complex and
inconstant, making it difficult to extract features sufficiently
discriminative} for aerial image representation. On the other
hand, the efficiency of the existing aerial image categorization
methods is far from practical {due to the
huge number of various components as well as their bilateral
relationships.} Therefore, a discriminative and concise aerial
image representation has become
increasingly imperative for a successful categorization system.\\
\indent In the literature of designing discriminative image
representations for visual recognition, many features have been
proposed. {They can be categorized into two groups: global
features} and local features. Global features, such as histograms,
eigenspace \cite{eigenspace}, and skeletal
shape~\cite{skeletalsharp}, generalize the entire image with a
single vector and are standard for statistic models like SVM.
However, global features are sensitive to occlusion and clutter.
{Besides, these representations typically rely on a preliminary
segmentation of objects in images.} These two limitations result
in unstable categorization performance. {Different from global
features, local features are developed to increase the
discrimination, such as scale invariant feature transform
(SIFT)~\cite{sift}.} Each local feature describes a localized
image region and is calculated around the interest points. Thus,
they are robust to partial occlusion and clutter. {To take
advantage of this property, local
features~\cite{handwritten,parsing,hierarchical} (\textit{e.g.},
junction~\cite{junction}, gradient~\cite{gradient},
contour,~\textit{etc}) are widely used for aerial image parsing
recently.} However, when employing local features for image
categorization, different images typically contain different
numbers of local features. {That is, it is difficult to integrate
the local features within an image for the standard classifiers.}
In many cases, they are integrated into an orderless
bag-of-features as global representation, thereby the similarity
between images is determined by the orderless bag-of-features. {It
is worth emphasizing that as a non-structural representation, the
bags-of-features representation ignores} the geometric property of
an image (\textit{i.e.}, the spatial distribution of the local
image patches), which prevents it from being highly
discriminative. Given the zebra skin and the chessboard skin,
their bag-of-features representations are similar. That is to say,
{the bag-of-features representation is not sufficiently
descriptive to distinguish the zebra and the chessboard, although
the geometric
properties of the two images are significantly different.\\
\indent {In order to encode image geometric proprieties into a
categorization model, several image geometric features} have been
proposed. In~\cite{beyond}, the spatial pyramid matching kernel is
obtained by clustering the local features into a few geometric
types. {However, the spatial pyramid matching kernel is not
flexible enough, since it highly depends on the human prior
knowledge.} RGB-domain spin image~\cite{spin} describes the
spatial context by exploring the chain structure of pixels in each
RGB channel. {However,} the chain structure usually fails to
describe the spatial context with complicated structures. Walk
kernel~\cite{walk_kernel} is proposed to capture the walk
structures among image local features. {However, the unavoidable
totter phenomenon (\textit{i.e.}, one vertex may occur several
times in a walk) brings noise and hence limiting its
discrimination.} To obtain a better discrimination, parameters are
provided to tune the length of the chain~\cite{spin} or
walk~\cite{walk_kernel}. This operation leads to very redundant
structures. {Both the time consumption and the memory cost
increase remarkably as the structure number goes up.} Therefore, a
concise image structure representation is desired for accurate
aerial image categorization. Recently, many graph-based models are
applied in intelligence systems and multimedia. They can be used
as geometric image descriptors~\cite{zhang1,zhang2,zhang3,zhang4}
to enhance image categorization. Besides, these methods can be
used as image high-order potential descriptors of
superpixels~\cite{zhang5,zhang6,zhang7,zhang8,zhang9}. Further,
graph-based descriptors can be used as a general image aesthetic
descriptors to improve image aesthetics ranking, photo retargeting
and cropping~\cite{zhang10,zhang11,zhang12,zhang13}.\\
\indent In this paper, {we propose a novel aerial image
categorization system, which enables the exploration of the
geometric property embedded in  local features}. An aerial image
is represented by a graph, since graph is a natural and
descriptive tool to express the complicated relationships among
objects. By defining region connected graph (RCG), we decompose an
aerial image into a set of discriminative subgraphs. To capture
discriminative relationships among RCGs, a structure refinement
strategy is carried out to select highly discriminative and low
redundant structures. {Based on the refined structures, we extract
sub-RCGs accordingly and all the sub-RCGs from an aerial image}
form the discriminative spatial context. Finally, a quantization
{operation transforms the} discriminative spatial context into a
feature vector
for categorization.\\
\indent The major contributions of this paper are as follows: 1)
region connected graph (RCG), a graph-based representation that
describes the local patches and their topology for an areal image;
2) a structure {refinement algorithm that
selects highly} discriminative and low redundant structures among
the training RCGs; and 3) an efficient isomorphism
{subgraph extraction component} that
acquires the corresponding sub-RCGs.

\section{Region Connected Graph(RCG)}
An aerial image usually contains millions of pixels. If we treat
each pixel as a local feature, highly computational complexity
will make aerial image recognition intractable.
{Fortunately, an aerial image can be
represented by a collection of clusters because pixels are usually
highly correlated with their neighboring ones.} Each cluster
consists of neighboring pixels with consistent color intensities.
Thus, given an aerial image, we can represent it by a set of
regions instead of millions of pixels. {The
neighboring relationships between regions define the spatial
context of an aerial image.} Naturally, we can model this
representation as a labeled graph. {The
labels denote the local features of each region and each edge
connects pairwise neighboring regions.} In our work, we call this
representation region connected graph (RCG).\\
\begin{figure}[h]\centering
\includegraphics[scale=0.62]{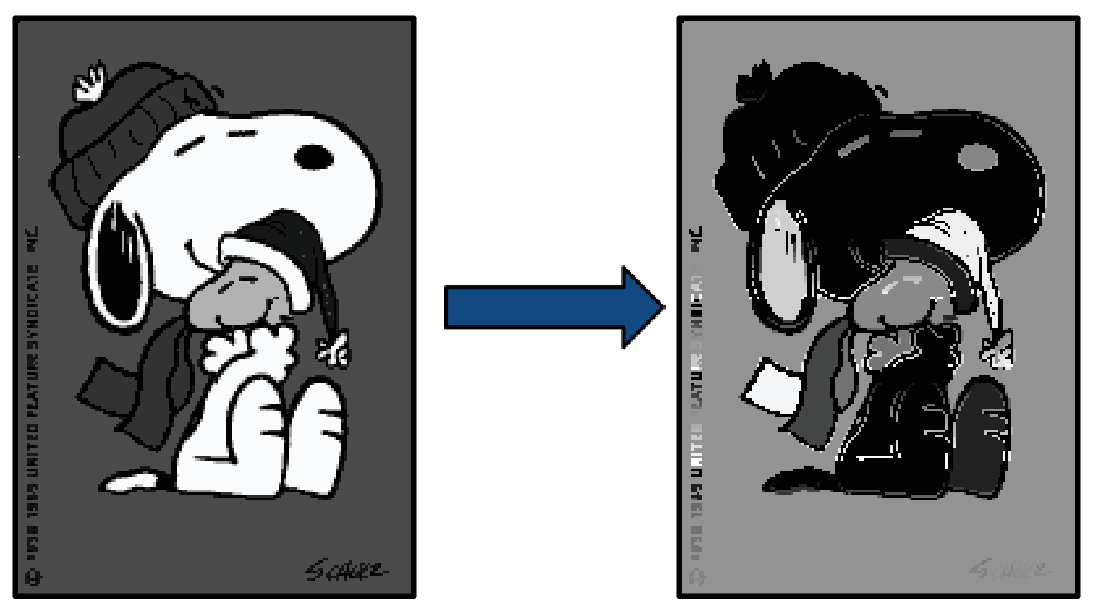}
\caption{From pixel clusters (left) to singly connected regions
(right).} \label{fig2}
\end{figure} \indent To obtain {the RCG} from an aerial image,
a segmentation algorithm (\textit{i.e.}, fuzzy
clustering~\cite{fuzzy} in our implementation) groups pixels into
different clusters according to their color intensity. Note that
the pixels in the same cluster are unnecessarily spatially
neighboring. As shown in Fig.~\ref{fig2}, we use different
grayscale values to identify different clusters. Pixels in the
face and the lower half of the Snoopy's body are grouped into the
same cluster. However, it is more reasonable if
{they are categorized into} different
groups, since the face {and the lower half
of Snoopy are spatially isolated. To this end,} a region growing
algorithm~\cite{ip_matlab} is employed to divide an image into
regions iteratively. In each iteration, the region growing
algorithm initializes the current region with a random pixel.
{It continues adding the spatially
neighboring pixels into this region if the current pixel and the
existing pixels come from the same cluster.} The iteration
terminates if the entire pixels are considered.
{The clustering
result is shown on the right of Fig.~\ref{fig2}.}\\
\begin{figure}[h]
\includegraphics[scale=0.5]{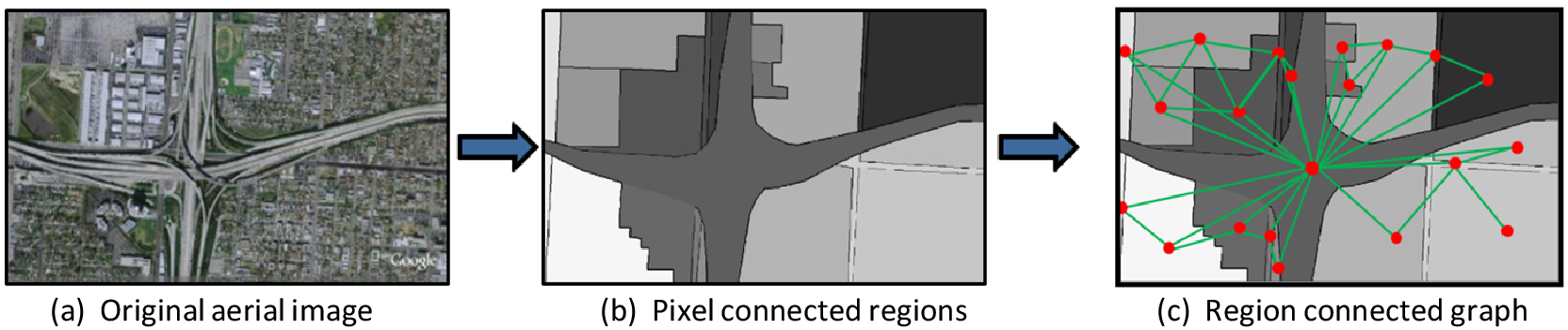}
\caption{The flowchart from an aerial image to its RCG.}
\label{fig3}
\end{figure} \indent On the basis of the singly connected regions,
the RCG of an aerial image can be obtained as shown in
Fig.~\ref{fig3}. Given an aerial image $I$ (Fig.~\ref{fig3}(a)),
we segment it into $K$ singly connected regions $R=\{r^i
\}_{i=1\cdots K}$ (Fig.~\ref{fig3}(b)).
{Then, each singly connected region is
treated as a vertex $v_i$ (the red solid point),} and the
relationship between spatially neighboring vertices is linked by
an edge $e_j$ (the green line). Finally, denoting $V=\cup v_i $ as
a collection vertices $v_i$ and $E=\cup e_j$ a set of edges $e_j$,
we define $\mathcal{G}=(V,E)$ as an RCG, where $V$ is a set of
{singly connected regions and $E$ is a set
of spatially neighboring} relationships (Fig.~\ref{fig3}(c)). Let
$|G|$ denote the number of vertices in RCG $G$. The number of
neighbors of a vertex is called the vertex degree.
{A useful attribute of RCG is that its
vertex degree is upper bounded. That is to say, each region has a
limited number of neighbors. It is observed that} the average
vertex degree of each RCG is less than four and the maximum vertex
degree is no more than 15.

\section{Discriminative Structures Selection}
It is natural to recognize {an aerial image
by matching its RCG to a labeled one.} However, as proved in
\cite{isomorphism}, given a pair of graphs, it is NP-hard to
determine whether they have the same structure. That means it is
intractable to {compare pairwise} RCGs
directly. Alternatively, we represent {an
aerial image by a set of sub-RCGs} $\{G_{sub}^k\}_{k=1\cdots N}$,
where $\cup_{k=1}^N G_{sub}^k=G$. Thereby, the aerial image
categorization can be conducted by matching its sub-RCGs to those
of the labeled aerial images. {Noticeably,
the RCG} of an aerial image may contain tens to hundreds of
vertices. Given $n$ vertices in an RCG, there will be
$N=2^{\frac{(n+1)*n}{2}}$ different sub-RCGs, which makes it
{impractical to} represent an aerial image
by enumerating all its sub-RCGs (Fig.~\ref{fig4}(a)). Toward a
discriminative and concise representation for aerial image
recognition, only sub-RCGs with highly discriminative and low
redundant structures should be {selected for
aerial image categorization (Fig.~\ref{fig4}(b)).}
\begin{figure}[h]
\includegraphics[scale=0.51]{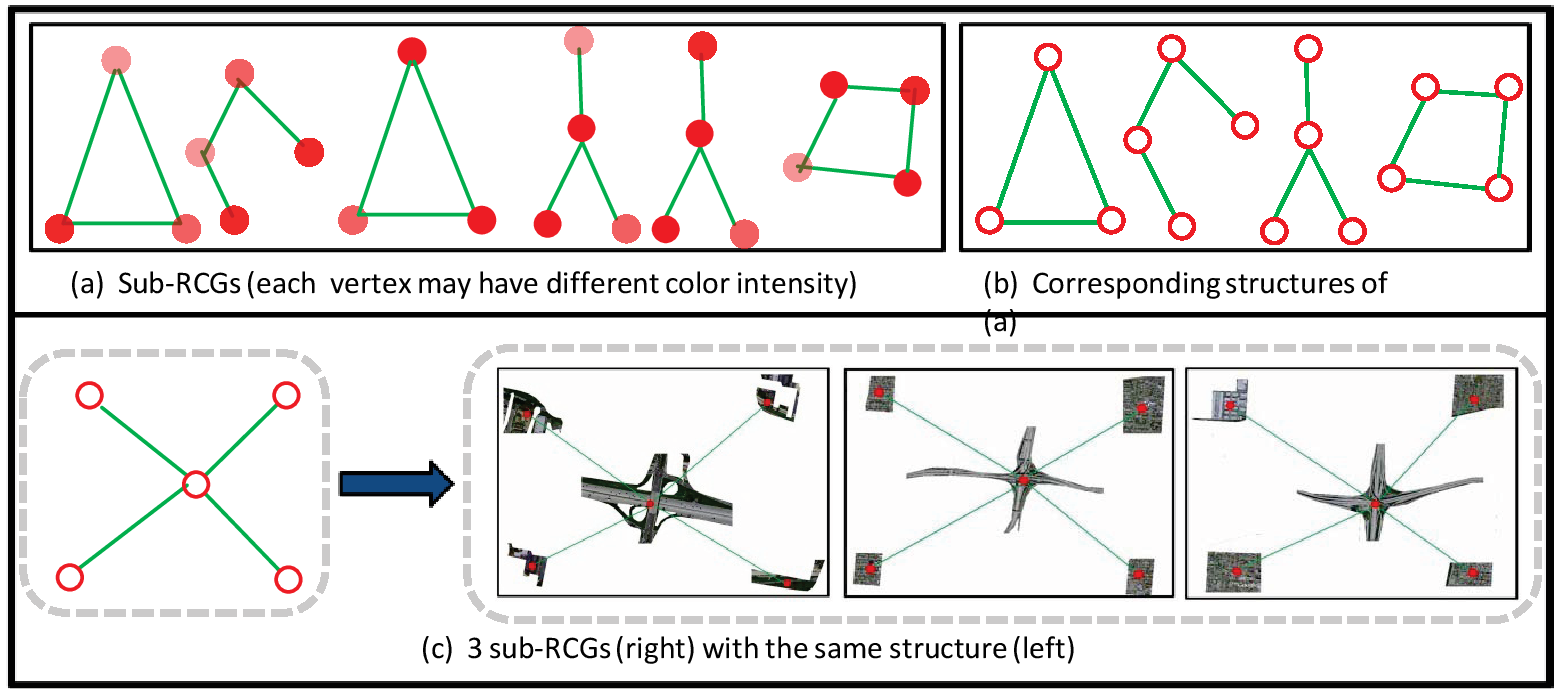}
\caption{{An example of sub-RCGs and their
structures.}} \label{fig4}
\end{figure}
\subsection{Frequent Structures Mining}
Each sub-RCG reflects the structure of a subset of connected
vertices in the RCG. In other words, a sub-RCG models the spatial
context of an aerial image. {Different types
of aerial image are with different spatial context, so do the
structures of sub-RCGs.} It is natural to use the structure of
sub-RCG to determine the aerial image type. For instance, as shown
in Fig.~\ref{fig4}(c), {all the three
sub-RCGs share the same structure but slightly different color
intensity} distributions. However, it is impractical to enumerate
all the possible sub-RCGs. Moreover, only those frequently
{occurred sub-RCGs contribute to the
recognition task} while the others are redundant. Motivated by
these, we have to select the frequent
{structures.}\\
\indent In our implementation, an efficient
{frequent subgraph discovery algorithm
called FSG~\cite{fsg}} is employed. It is noticeable that the
vertex value of sub-RCGs might be different though they share the
same structure. This prevents us from mining the frequent
structures accurately. {Therefore, we ignore
the difference of vertex values. In particular,} given a sub-RCG,
its structure $S$ is obtained by setting the vertex labels of the
sub-RCG to a same value,~\textit{e.g.}, one.\\
\indent FSG accumulates the times of happening for each structure.
{It outputs the probabilities of all the
structures in the training RCGs, implying that the structure is
unnecessarily existing in all the training RCGs.} A probability
$p(S)$ represents the frequency of $S$. As the number of original
candidate structures is exponential, only the structure whose
probability is higher than a threshold is output as a frequent
one. {Therefore, the number of frequent
structures is greatly reduced greatly.}
\subsection{Measures for Structure Selection}
The number of frequent structures is still too large (typically
100$\sim$300) though it is much smaller than that of the candidate
structures. {In addition, a structure with
high frequency may not be highly discriminative.} Thus, we carry
out a further selection among the frequent structures
{to preserve only the highly discriminative
and low redundant ones.} We first define a distance to describe
the similarity between sub-RCGs ($G_{sub}$ and $G_{sub}')$ with
the same size:
\begin{equation}
d(G_{sub},G_{sub}')=\sum_{v_r\subseteq G_{sub}\wedge v_r'\subseteq
G_{sub}'}{\parallel f(v_r)-f(v_r')\parallel}^2 \label{eq_1}
\end{equation}
where $v_r$ is $r$-th vertex of $G_{sub}$ and $f(\cdot)$  the
local regions' feature vector. $||\cdot||$  is the Euclidean norm.
More specifically, for structure $S$ and $S'$ in $G$ and $G'$
respectively, if $|S|=|S'|$, we define the structure distance
between $G$ and $G'$ as follows:
\begin{equation}
 d_e\left(S_G,S_{(G')}'\right)=\varphi\cdot \hspace{-10pt}\sum_{G_{sub}(S)\subseteq G}\cdot \sum_{G_{sub}' (S')\subseteq G'}
 \hspace{-10pt}d(G_{sub} (S),G_{sub}'(S'))
\label{eq_2}
\end{equation}
where $G_{sub}(S_{(\cdot)})$ is the sub-RCG corresponding to
$S_{(\cdot)}$. {$\varphi$ is a factor that
normalize $d_e$ to $[0, 1]$ and it is not a tuning parameter. That
is, $\varphi=\frac{1} {(|G_{sub}|\cdot|G_{sub}'|}$, where
$|G_{sub}|$ and $|G_{sub}'|$ denote the number of sub-RCGs in RCG
$G$ and $G'$, respectively.} By extending Eq.(~\ref{eq_2}) to the
situations when $\|S\|\neq\|S'\|$, we define a more generic form
of the structure distance between $G$ and $G'$.
{It is based on the probability $p(S)$ by
taking into account of different situations.}
\begin{eqnarray}
d\left(S_G,S_{G'}'\right)=\left\{ \begin{array}{ll} p(S)*p(S')*d_e\left(S_G,S_{(G')}'\right)\\
\hspace{10pt}\textrm{if $|S|=|S'|$ and $G_{sub}\neq \emptyset  \wedge G_{sub}'\neq \emptyset $}\\
p(S)*p(S')*\sum_i d_e\left(S_G,C_i(S_{G'},S_G)\right) \\
\hspace{10pt}\textrm{if $|S|<|S'|$ and $G_{sub}\neq \emptyset  \wedge G_{sub}'\neq \emptyset $}\\
p(S)*p(S')*\sum_i d_e\left(S_G',C_i(S_{G},S_G')\right) \\
\hspace{10pt}\textrm{if $|S|>|S'|$ and $G_{sub}\neq \emptyset  \wedge G_{sub}'\neq \emptyset $}\\
(1-p(S))*(1-p(S'))\\
\hspace{10pt}\textrm{if $G_{sub}= \emptyset  \wedge G_{sub}'= \emptyset $}\\
p(S)+p(S')-2p(S)*p(S')\\
\hspace{10pt}\textrm{otherwise}\\
\end{array}\right.\label{eq_3}
\end{eqnarray}
\indent The probability for structure $S$ existing in $G$ is
denoted by $p(S)$. It is straightforward to obtain the first line
of Eq.(\ref{eq_3}) by multiplying $p(S)*p(S^{'})$ with the
structure distance $d_e(S,S')$ wherein $p(S)*p(S')$ denotes the
probability for $S$ existing in $G$ and $S_2$ existing in $G'$.
{This is similar to the second line and the
third line of Eq.(\ref{eq_3}).} As $S$ is a subset of $S'$ when
$|S|<|S'|$, the function $C(S,S')$ outputs the enumerated
structures with the same size to $S'$ in $S$ by FSG~\cite{fsg} in
the second line of Eq.(\ref{eq_3}), and vice versa in the third
line. $(1-p(S))*(1-p(S'))$ denotes the probability for neither $S$
existing in $G$ nor $S'$ existing in $G'$. An
$p(S)+p(S')-2p(S)*p(S')$ in the last line is the probability for
either $S$ existing in either $G$ or $S'$ existing
in $G'$. $d\left(S_G,S_{G'}' \right)\in [0,1]$.\\
\indent Based on the structure distance
$d\left(S_G,S_{G'}'\right)$ between $G$ and $G'$,
{measure of structure discrimination(MSD),
is defined for structure's discrimination.} Inspired by the
definition of discriminative ability in LDA~\cite{klda}, MSD
computes the distance ratio between RCGs with different labels and
those with same labels:
\begin{equation}
M_{sd}(S)=\frac{D_{S}^{b}}{D_{S}^{w}}= \frac{\sum_{G}\sum_{G'}
d\left(S_G,S_{G'}'\right)*\sigma} {\sum_{G}\sum_{G'}
d\left(S_G,S_{G'}'\right)*\sigma'} \label{eq_4}
\end{equation}
$\sigma$ and $\sigma'$ {are functions
indicating whether $G$ and $G'$ are} belong to the same class. If
$G$ and $G'$ belong to different classes, $\sigma=1,\sigma'=0$,
otherwise $\sigma=1,\sigma'=0$\footnote{Pairwise RCGs $G$ and $G'$
belonging to the same class means that their corresponding aerial
images belong to the same class. Similarly, two RCGs $G$ and $G'$
belonging to different classes means that their corresponding
aerial images belong to different classes. }. A larger $M_{sd}(S)$
means a more discriminative ability of structure $S$. However, a
structure set with {high discrimination
doesn't} mean it is a concise one. Aiming at a concise set of
structures, it is necessary to make further structure selection.
{Motivated by the fact that high correlation
leads to high redundancy~\cite{speech}, we believe that one of the
two structures should be removed if two structures are highly
correlated.} In order to calculate the correlation between
structures, an approach to quantize the redundancy between
structures, {called measure of structures
correlation (MSC), is defined based on the distance between
structures:}
\begin{equation}
M_{sc}(S,S')=\frac{\sum_{G}\sum_{G'} d\left(S_G,S_{G'}'\right)}
{\sum_{G}\sum_{G'} d\left(S_G,S_{G'}\right)+\sum_{G}\sum_{G'}
d\left(S_G',S_{G'}'\right)} \label{eq_5}
\end{equation}
where the denominator functions as a normalization step. A larger
$M_{sc}(S,S')$ leads to a lower correlation between structure $S$
and $S'$, and vice versa. Eq.(\ref{eq_5}) also can be explained by
analogy with the three vertices of a triangle in Fig.~\ref{fig_5}.
$\sum_{G}\sum_{G'} d\left(S_G,S_{G'}'\right)$, $\sum_{G}\sum_{G'}
d\left(S_G,S_{G'}\right)$ and $\sum_{G}\sum_{G'}
d\left(S_G',S_{G'}'\right)$ act as the distance between the three
vertices. {When
$\sum_{G}\sum_{G'}d\left(S_G,S_{G'}'\right)$ becomes larger, the
correlation between $S$ and $S'$ becomes lower
(Fig.~\ref{fig_5}(c)), and vice versa (Fig.~\ref{fig_5}(b)).}
\begin{figure}[h]
\includegraphics[scale=0.44]{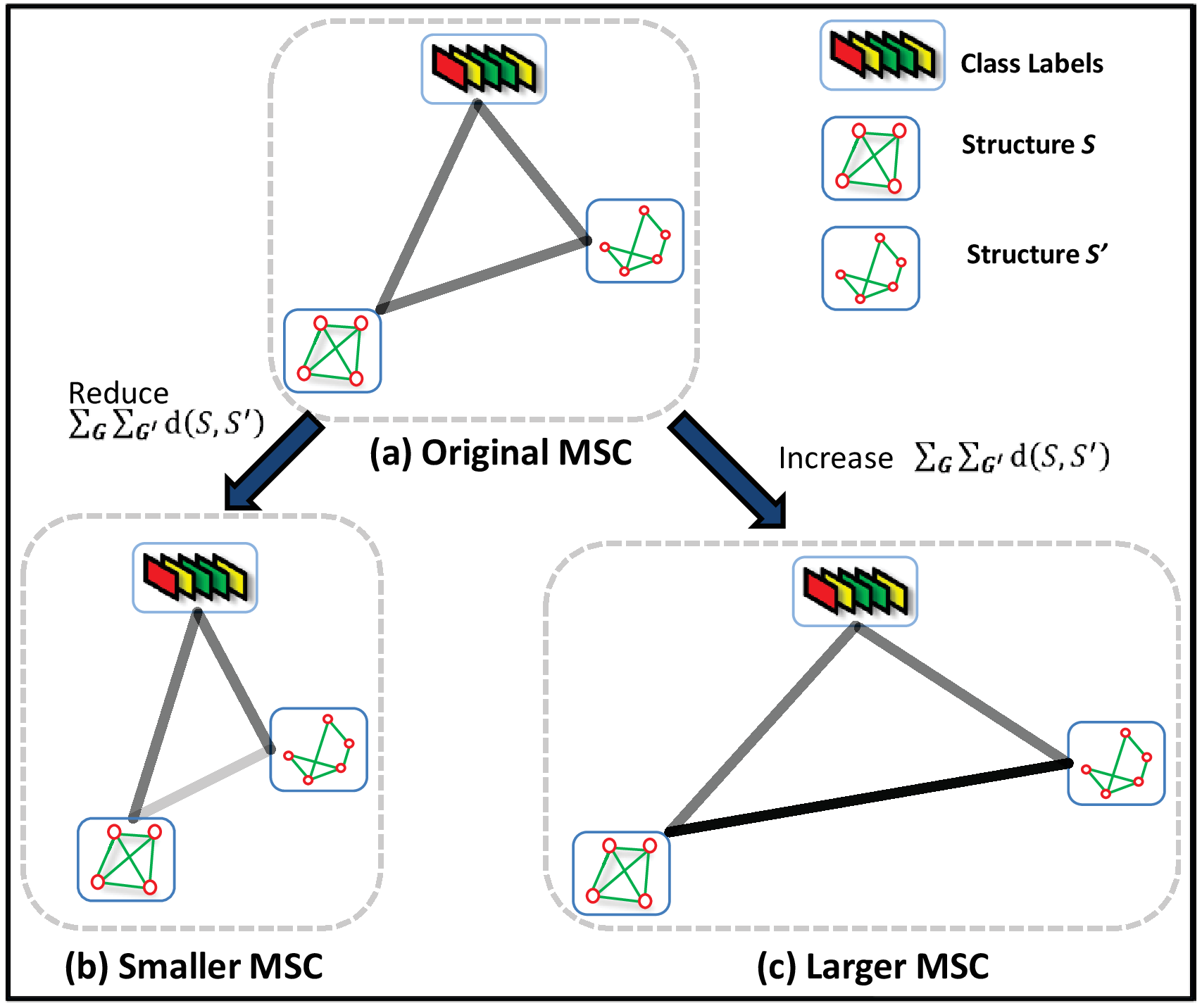}
\caption{A visual explanation of the correlation between
structures.} \label{fig_5}
\end{figure}\subsection{MSD and MSC based Structure Refinement}
Based on the two structure measures MSD and MSC, we construct a
novel concise and discriminative structure refinement algorithm.
{The stepwise operations of the proposed structure selection are
illustrated in and Algorithm~\ref{tab_1} respectively.} The
algorithm can be divided into two steps. First, the MSD values of
all the candidate structures are computed and sorted in descending
order. Candidate structure whose MSD value is higher than a
threshold will be preserved initially into the list $L_{final}$.
Second, the MSC value between each pair of preserved structures
{is computed to evaluate their redundancy.} The removal of
redundant structures is carried out iteratively. During the first
round of iteration, we specify the preserved structure with the
largest MSD value as the final selected one. Then, we sort the MSC
values between the finally selected structure and the rest of the
preserved structures. The structure whose MSC value is higher than
a threshold will be removed. {The preserved structure list will be
updated accordingly.} After one round of iteration, we move to the
preserved structure with lower MSD value. The iteration terminates
when there is no structure next to $S_{final}$. The finally
preserved structures are deemed as the refined ones.\\
\begin{algorithm}\centering \caption{MSD\&MSC-based
Structure Refinement}
\begin{tabular}{l}
\textbf{Input}:
 $R\{S_1,S_2\cdots S_{N},C\}$ \hspace{10pt}  $//$training data set \\
~~~~$\delta_{sd},\delta_{sc} $  \hspace{52pt}$//$the threshold for MSD and MSC \\
\textbf{Output}: $R_{final}$     \hspace{65pt}//a set of refined
structures\\\hline
\hspace{0pt}for $i$~=~$1$ : $N$ \textbf{do begin} \hspace{20pt} $//$ step1\\
\hspace{10pt}calculate $M_{sd}$ for $S_i$; \\
\hspace{20pt}if$\left(M_{sd}(S_i)>\delta_{sd}\right)$\\
\hspace{30pt}preserve $S_i$ into $R^{'}$;\\
\hspace{30pt}order $R_{'}$ in descending $M_{sd}$ value;\\
\hspace{10pt}  \textbf{end};\\
\hspace{0pt}  $S_{final}=getFirstStructure(R^{'})$; \hspace{20pt} $//$ step2  \\
\hspace{10pt} \textbf{do begin} \\
\hspace{20pt} $S_{tmp1}=getNextStructure(R^{'},S_{final})$; \\
\hspace{20pt} \textbf{do begin} \hspace{30pt}   $//$ remove redundant structures \\
\hspace{30pt} $S_{tmp2}=S_{tmp1}$;\\
\hspace{30pt} if$\left(M_{sc}(S_{final},S_{tmp2})>\delta_{sc}\right)$\\
\hspace{40pt} remove $S_{tmp1}$ from $R^{'}$;\\
\hspace{40pt} $S_{tmp1}=getNextStructure(R^{'},S_{tmp2})$;\\
\hspace{30pt} else\\
\hspace{40pt}  $S_{tmp1}=getNextStructure(R^{'},S_{tmp1})$;\\
\hspace{20pt} \textbf{end until}$\left(S_{tmp1}==NULL\right)$;\\
\hspace{20pt} add $S_{final}$ to $R_{final}$;\\
\hspace{20pt} $S_{final}=getNextStructure(R^{'},S_{final})$;\\
\hspace{10pt} \textbf{end
until}$\left(S_{final}==NULL\right)$;\\\end{tabular} \label{tab_1}
\end{algorithm} \indent Denote $n$ as the number of training RCGs and $m$ as the number
of candidate structures, {we assume that the
structure distance between RCGs can be computed in constant time.
As the distance between RCGs is required} for calculating MSD and
MSC, the computational cost of calculating MSD and MSC are both
$\mathcal{O}(n^2)$. {As shown in
Algorithm~\ref{tab_1}, the structure refinement step contains a
double loop and the time complexity of each is
$\mathcal{O}(n^2*m)$.} Therefore, the time complexity of the whole
selection process is $\mathcal{O}(n^2*m^2)$.

\section{Geometric Discriminative Feature}
\subsection{Geometric Discriminative Feature Extraction}
As the {refined structures are both concise
and discriminative,} they are adopted to extract the geometric
discriminative features. Guided by the refined structures, we
extract sub-RCGs with the same structures and then use them as the
geometric discriminative features. As RCGs are
{low degree graphs (vertex degree less than
15),} the computational complexity is nearly linear
increasing with the number of vertices~\cite{walk_kernel}.\\
\indent {To achieve an efficient sub-RCG extraction process, we
propose} an algorithm to locate the sub-RCGs efficiently. Given a
refined structure $S$ and an RCG $G$, the proposed algorithm
outputs a collection of sub-RCGs with structure $S$. There are
three steps in the proposed geometric discriminative feature
extraction. First, the vertices {of $S$ are checked to determine}
whether $|S|\leq |G|$. {If $|S| \leq |G|$, then an iterative
process will be carried out.} Otherwise, the algorithm will
terminates. Next, for each vertex in $G$, we treat it as the
reference point and compare $S$ to the structures of its
correlated sub-RCGs. A depth-first-search strategy~\cite{dfs} is
employed for graph matching. Only the sub-RCGs with the same
structure to $S$ are {the preserved.} By traversing all the
vertices in RCG $G$, we perform the matching process and collect
all the qualified sub-RCGs. Finally, a collection of qualified
sub-RCGs are obtained

\subsection{Quantizing Sub-RCGs into Feature Vectors}
Given an aerial image, it can be represented by a set of sub-RCGs
as described above. It is worth {emphasizing
that the sub-RCGs are planar visual feature in $\mathbb{R}^2$.}
Conventional classifiers such as support vector machine
(SVM)~\cite{ksvm} can only handle 1-D vectors. Further, the number
the extracted sub-RCGs are different from one aerial image to
another. Therefore, it is {impractical for a
conventional classifier like SVM to carry out classification
directly.} To tackle this problem, a quantization method is
developed to convert each aerial image into a 1-D vector.\\
\begin{figure}[h]
\includegraphics[scale=0.55]{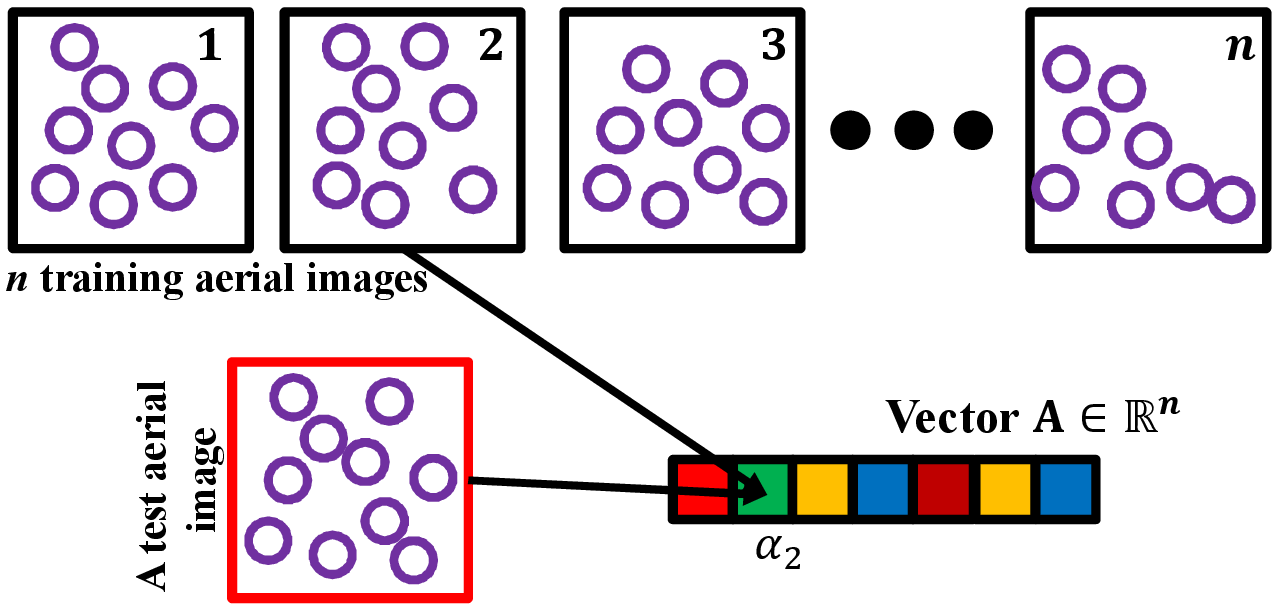}
\caption{An illustration of generating the feature vector for a
test aerial image. The blue circles in each aerial image denote
the sub-RCGs.} \label{fig_add1}
\end{figure}\indent The proposed quantization method is based on the distances
between the test aerial images and the training ones.
{The distance is computed using the
extracted geometric discriminative features.} Given an aerial
image, we first extract its geometric discriminative features,
each corresponding to a refined structure. Then. as shown in
Fig.~\ref{fig_add1}, an aerial image is encoded into a vector
$\mathbf{A}=[\alpha_1, \alpha_2\cdots,\alpha_n]^T$, where $n$ is
the number of training
aerial images and each element of $\alpha$ is computed as:\\
\begin{equation}
\alpha_i\propto \exp(-\lambda*\sum\nolimits_S d\left(S_G,S_{G^{'}})\right)\\
\label{eq_6} \end{equation} where $\lambda$ is a free parameter to
be tuned. {In our implementation, we fix
$\lambda$ to 0.5 by using cross validation.}

\section{System Overview}
Our aerial image categorization system can be divided into the
training and the test stages. In the training phase, structure
refinement for geometric discriminative feature extraction is
conducted. First, each aerial image is segmented into connected
regions for building the corresponding RCGs. {Then, a frequent
structure mining algorithm is employed to discover the highly
frequent structures in the training RCGs.} Next, MSD and MSC are
computed for each structure toward a concise set of structures.
Structure refinement is carried out to acquire the highly
discriminative and low redundant ones. Third, the geometric
{discriminative features are obtained by extracting the sub-RCGs}
corresponding to the refined structures. To convert the extracted
2-D geometric discriminative features into 1-D vectors, {a
quantization scheme computes the distance} between the given
aerial image and the training samples. Finally, we {train an SVM
classifier by the vectors from} the encoded training samples.\\
\indent The test phase {is illustrated on the right.} Given a test
aerial image, we obtain its RCG firstly. Then, the geometric
discriminative features are extracted to represent the given
aerial image. {Similarly,  a quantization operation is carried out
to convert the aerial image into a vector using the geometric
discriminative features.} This vector is fed into the trained SVM
for aerial image categorization.

\section{Experiments and Results Analysis}
Experiments are carried out on two data sets.
{The first data set contains the aerial
images from the Lotus Hill (LHI) data set~\cite{lotus}.} It
consists of five categories where each category contains 20 aerial
images. Each image is associated with a standard segmentation map.
The second data set is our own complied data set and it includes
aerial images from ten categories . {The
whole data set contains 2,096 aerial images crawled from the
Google Earth.} The experimental system is equipped with an Intel
E8500 CPU and 4GB RAM. {All the algorithms
are implemented on the Matlab platform.}
\subsection{Comparative Study}
In our experiment, the validation of the proposed geometric
discriminative feature {is conducted on both } the LHI and our own
data sets. We compare our geometric discriminative feature {with
several representative discriminative visual}
features,~\textit{i.e.}, the global RGB histogram, the
intensity-domain spin images~\cite{spin}, the walk/tree
kernel~\cite{walk_kernel}, the sparse coding spatial pyramid
matching (SC-SPM)~\cite{scSPM}, the locality-constrained spatial
pyramid matching (LLC-SPM)~\cite{llcSPM}, and the object
bank~\cite{ob}. As the spatial pyramid matching
kernel~\cite{beyond} heavily relies on the prior knowledge, we do
not employ it for comparison. {In our implementation, the
geometric discriminative features are extracted to encode both the
color intensity distribution and the spatial property.} In each
segmented region, a 4096-dimensional RCB-histogram is extracted as
its representation. A few example
aerial images and their geometric discriminative features are presented.\\
\begin{table*}\tiny \centering
\caption{{Recognition rate with standard
deviation on our own data set}(the experiment was repeated 10
times; HC is the HOG+color moment with a 1024-sized codebook; the
number in each bracket denotes the codebook size; and LR2 and LRG
are different regularizers as described in~\cite{ob})}
\begin{tabular}{|l|c|c|c|c|c|c|c|c|c|c|}\hline
Category &Walk kernel &Tree kernel& SPM(200) &SC-SPM(256) &
LLC-SPM(256) &OB-SPM(LR1)  &SPM(400) &SC-SPM(512)\\\hline Airport
&0.882$\pm$0.023 &0.901$\pm$0.032 &0.723$\pm$0.017
&0.721$\pm$0.026 &0.723$\pm$0.017 &0.799$\pm$0.021
&0.811$\pm$0.043 &0.843$\pm$0.021 \\\hline Commer.
&0.545$\pm$0.034 &0.532$\pm$0.012 &0.441$\pm$0.023
&0.443$\pm$0.031 &0.334$\pm$0.027 &0.517$\pm$0.036
&0.521$\pm$0.022&0.456$\pm$0.012 \\\hline Indust.& 0.642$\pm$0.021
&0.611$\pm$0.032 &0.521$\pm$0.021 &0.499$\pm$0.041
&0.413$\pm$0.015 &0.512$\pm$0.056 &0.454$\pm$0.033&0.576$\pm$0.018
\\\hline Inter. & 0.645$\pm$0.067 &0.685$\pm$0.011
&0.611$\pm$0.018 &0.643$\pm$0.023 &0.322$\pm$0.031
&0.675$\pm$0.034 &0.674$\pm$0.026&0.634$\pm$0.011\\\hline Park.  &
0.523$\pm$0.039 &0.487$\pm$0.017 &0.443$\pm$0.011 &0.512$\pm$0.037
&0.412$\pm$0.021 &0.536$\pm$0.012
&0.512$\pm$0.057&0.496$\pm$0.025\\\hline Railway &0.556$\pm$0.076
&0.578$\pm$0.056 &0.502$\pm$0.032 &0.511$\pm$0.022
&0.521$\pm$0.033 &0.514$\pm$0.013 &0.521$\pm$0.038
&0.596$\pm$0.052\\\hline Seaport& 0.859$\pm$0.051 &0.843$\pm$0.036
&0.774$\pm$0.021 &0.745$\pm$0.034 &0.721$\pm$0.034
&0.766$\pm$0.016 &0.632$\pm$0.043&0.814$\pm$0.009\\\hline Soccer
&0.646$\pm$0.021 &0.655$\pm$0.006 &0.576$\pm$0.021
&0.589$\pm$0.023 &0.578$\pm$0.023 &0.568$\pm$0.032
&0.521$\pm$0.045&0.624$\pm$0.032\\\hline Temple  &0.503$\pm$0.029
&0.454$\pm$0.031 &0.521$\pm$0.042 &0.567$\pm$0.038
&0.511$\pm$0.031 &0.603$\pm$0.021
&0.534$\pm$0.024&0.565$\pm$0.045\\\hline Univer. &0.241$\pm$0.045
&0.265$\pm$0.009 &0.289$\pm$0.017 &0.301$\pm$0.021
&0.223$\pm$0.044 &0.304$\pm$0.041
&0.498$\pm$0.03&0.321$\pm$0.012\\\hline Average &0.524$\pm$0.041
&0.601$\pm$0.024 &0.540$\pm$0.022 &0.553$\pm$0.030
&0.4770$\pm$0.033 &0.579$\pm$0.028
&0.568$\pm$0.037&0.593$\pm$0.024\\\hline\hline Category &LLC-SPM
(512) &OB-SPM (LRG) &SPM(800) &SC-SPM(1024)& LLC-SPM(1024)
&OB-SPM(LRG1)&   SPM(HC)  & SC-SPM(HC) \\\hline Airport
&0.801$\pm$0.021 &0.889$\pm$0.035  &0.799$\pm$0.033
&0.912$\pm$0.015 &0.899$\pm$0.019 &0.872$\pm$0.051 &
0.813$\pm$0.045   & 0.916$\pm$0.023 \\\hline Commer.
&0.567$\pm$0.034 &0.565$\pm$0.032 &0.512$\pm$0.032
&0.601$\pm$0.034 &0.521$\pm$0.021 &0.617$\pm$0.034 &
0.519$\pm$0.043   & 0.584$\pm$0.042 \\\hline Indust.&
0.521$\pm$0.025 &0.613$\pm$0.013 &0.585$\pm$0.043
&0.557$\pm$0.032&0.593$\pm$0.019 &0.576$\pm$0.054 &0.598$\pm$0.058
& 0.564$\pm$0.039 \\\hline Inter. & 0.766$\pm$0.036
&0.705$\pm$0.015  &0.644$\pm$0.022
&0.788$\pm$0.014&0.622$\pm$0.035 &0.676$\pm$0.013 &
0.668$\pm$0.041   & 0.791$\pm$0.019 \\\hline Park.  &
0.489$\pm$0.032 &0.486$\pm$0.016  &0.503$\pm$0.043
&0.489$\pm$0.043&0.489$\pm$0.055 &0.512$\pm$0.009 &
0.511$\pm$0.057  &  0.487$\pm$0.025 \\\hline Railway
&0.553$\pm$0.042 &0.532$\pm$0.053 &0.602$\pm$0.017
&0.601$\pm$0.037&0.599$\pm$0.009 &0.589$\pm$0.010 &
0.614$\pm$0.026   & 0.609$\pm$0.044 \\\hline Seaport&
0.751$\pm$0.036 &0.779$\pm$0.045  &0.815$\pm$0.031
&0.745$\pm$0.034&0.798$\pm$0.032 &0.811$\pm$0.013 &
0.822$\pm$0.039   & 0.751$\pm$0.039 \\\hline Soccer
&0.625$\pm$0.026 &0.646$\pm$0.014  &0.634$\pm$0.028
&0.689$\pm$0.036&0.655$\pm$0.014 &0.668$\pm$0.043 &
0.643$\pm$0.037   & 0.693$\pm$0.045 \\\hline Temple
&0.567$\pm$0.024 &0.587$\pm$0.027  &0.577$\pm$0.041
&0.689$\pm$0.027&0.556$\pm$0.032 &0.612$\pm$0.025 &
0.587$\pm$0.046   & 0.649$\pm$0.034 \\\hline Univer.
&0.409$\pm$0.042 &0.389$\pm$0.018  &0.311$\pm$0.013
&0.582$\pm$0.035&0.281$\pm$0.042 &0.304$\pm$0.011 &
0.324$\pm$0.031   & 0.537$\pm$0.033 \\\hline Average
&0.605$\pm$0.032 &0.620$\pm$0.027  &0.606$\pm$0.029
&0.654$\pm$0.033&0.600$\pm$0.027 &0.636$\pm$0.025 &
0.610$\pm$0.042   & 0.658$\pm$0.032 \\\hline Category   &
LLC-SPM(HC) &Our proposed method &&&&&&\\\hline Airport  &
0.904$\pm$0.031 &0.864$\pm$0.051&&&&&&\\\hline Commer.  &
0.534$\pm$0.029&0.677$\pm$0.024&&&&&&\\\hline Indust.   &
0.598$\pm$0.023&0.555$\pm$0.034&&&&&&\\\hline Inter.   &
0.634$\pm$0.046&0.812$\pm$0.021&&&&&&\\\hline Park.    &
0.493$\pm$0.064&0.501$\pm$0.061&&&&&&\\\hline Railway   &
0.604$\pm$0.005&0.606$\pm$0.033&&&&&&\\\hline Seaport  &
0.803$\pm$0.046&0.771$\pm$0.025&&&&&&\\\hline Soccer   &
0.659$\pm$0.026&0.663$\pm$0.065&&&&&&\\\hline Temple   &
0.574$\pm$0.041&0.665$\pm$0.019&&&&&&\\\hline Univer.  &
0.287$\pm$0.049&0.551$\pm$0.034&&&&&&\\\hline Average  &
0.609$\pm$0.036&0.667$\pm$0.037&&&&&&\\\hline
\end{tabular}\label{tab1a}
\end{table*}
 \indent First, we present a set of discovered
discriminative subgraphs. From a horizontal glance, we can roughly
discriminate aerial images from the five categories, especially
for the intersections and the marines. This demonstrates {the
necessity to exploit the relationships among aerial image patches}
for categorization.\\
\indent Further, to make comparison among the global histogram,
the spin images, the walk kernel, and the proposed geometric
discriminative feature, {we select half of
the images for training} and leave the rest for testing. As shown
in Table~\ref{tab1a}, the proposed {feature
achieves} the best accuracy on average.
\subsection{Discussion on different parameter settings}
We notice that the {influence of
segmentation operation in the RCG construction is unnegligible.}
To evaluate the performance under different segmentation settings
(\textit{i.e.}, the number of singly connected regions), we
perform aerial {image recognition on the LHI
data set, since the off-the-shelf segmentation benchmark} is
suitable to make a fair comparison.\\
\indent Different segmentation settings are employed in our
evaluation,{~\textit{i.e.}, deficient segmentation and over
segmentation.} The MSD values of each aerial image corresponding
to different segmentation settings are computed. {We observed that
the benchmark segmentation setting achieves} the largest MSD value
6.3, while the deficient segmentation and over segmentation gain
4.9 and 5.7, respectively. {Comparatively, more regions are
obtained} in overly segmentations, which means it is rarer for one
region to span several objects. Therefore, when building an RCG
{by overly segmented regions,} fewer discriminative objects are
neglected. {Further, it is unavoidable that the unsupervised
clustering is less accurate than
the benchmark segmentation.}\\
\begin{table} \centering \caption{Recognition accuracy {under
different segmentation schemes}}
\begin{tabular}{|c|c|c|c|c|}\hline
Category &Bench. &Defic.   &Overly &Mulit.\\\hline Intersection&
0.8 &0.3 &0.8 &0.8\\\hline Marine & 0.4&0.8 &0.8 &0.9\\\hline
Parking & 0.9& 0.5&0.6 &0.6
\\\hline Residental & 0.5 &0.7&0.6 &0.7\\\hline School &0.6
&0.3&0.3&0.6
\\\hline Average rate &0.64 &0.54 &0.62 &0.72\\\hline Total
topology \# &73 &125&177 &143\\\hline Selected structure \# &8 &8
&8 &8\\\hline Average RAG edge \# &37 &26  &57 &41\\\hline Average
RAG vertex \# &19 &16 &31 &19\\\hline
\end{tabular} \label{tab2a}\end{table}
\indent We compare {the categorization
accuracy under} the benchmark segmentation, the over segmentation
and the deficient segmentation. {As shown in
Table~\ref{tab2a},} over segmentation obtains 2$\%$ lower accuracy
than that of the benchmark segmentation on average. Deficient
segmentation performs worse than over segmentation by providing
the lowest accuracy. The overall recognition result is consistent
to what the MSD reflects.\\
\begin{figure}[h]
\includegraphics[scale=0.61]{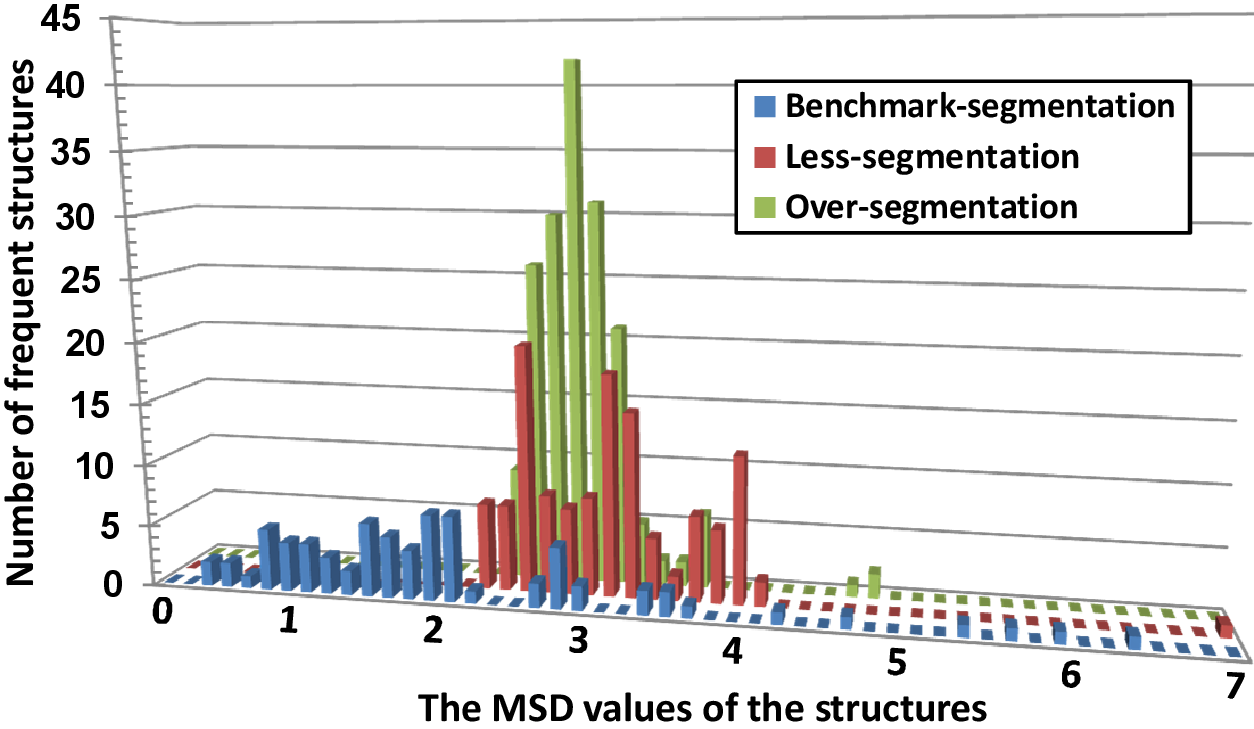}
\caption{{The discrimination of the frequent
structures under} different segmentation schemes.} \label{fig11}
\end{figure}
\indent In the structure selection stage, both the threshold of
MSD and MSC influence the obtained structures. Toward an easy
parameter tuning process, we set the threshold of MSD to a small
value, which allows a large number of candidate structures to be
qualified. Then, we tune of threshold of MSC to carefully remove
those redundant structures. As shown in Fig.~\ref{fig_add2}, we
set the threshold of MSD to 0.1 and tune the threshold of MSC. It
is observed that the categorization accuracy increases and then
becomes the threshold of MSC reaches 0.65. Thus, we set the
thresholds of MSD and MSC to 0.1 and 0.65 in our implementation.
\begin{figure}[h]
\includegraphics[scale=0.53]{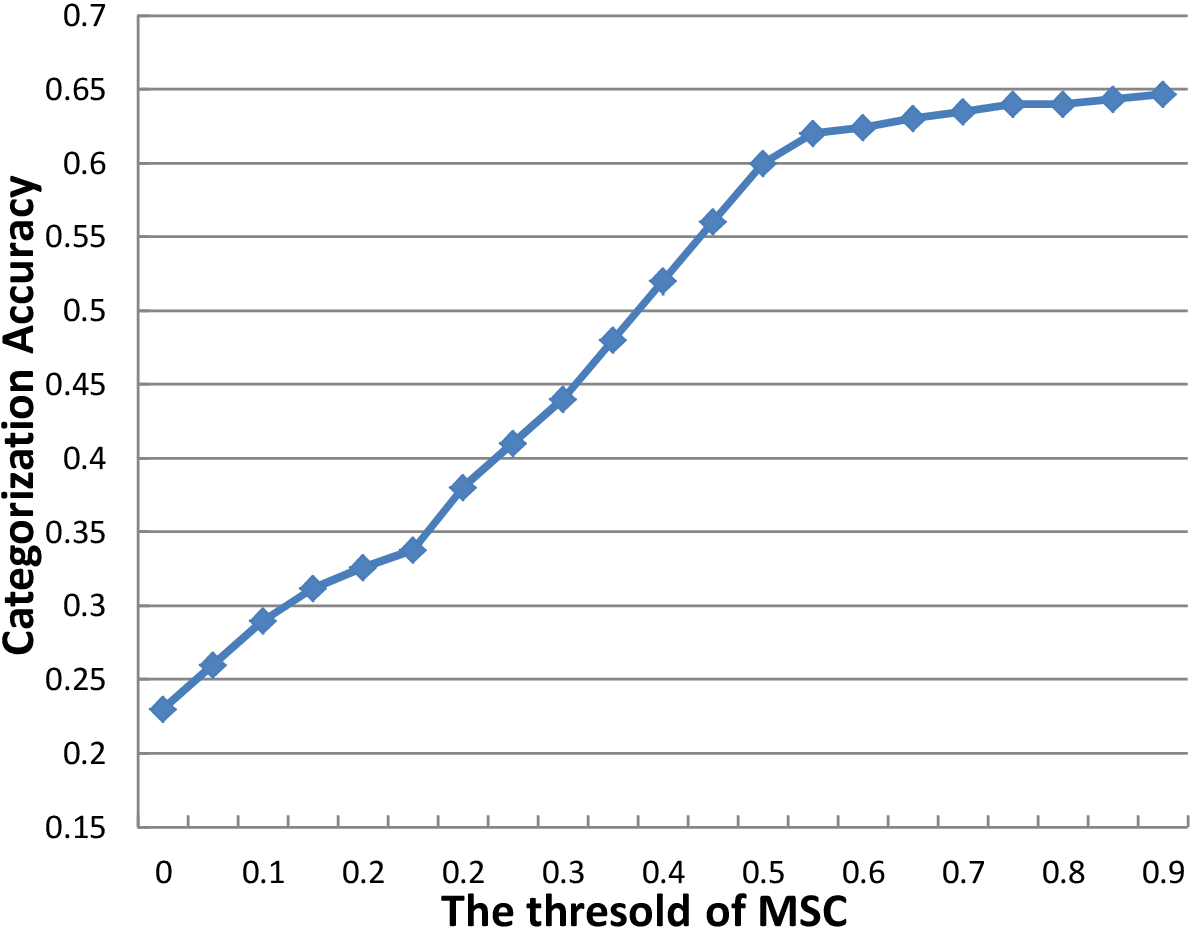}
\caption{The categorization performance under different MSC
thresholds.} \label{fig_add2}
\end{figure}
\subsection{The compilation of our aerial image data set}
We {compiled our data set by searching
aerial images from the Google Earth.} The whole data set contains
2,096 aerial images from ten categories. Since the aerial images
from cities are usually clearer than those from the remote areas,
we collected most of our images from metropolis, such as New York,
Tokyo and Beijing. Due to {the various
difficulties to crawl images} from different categories, the
number of images in each category varies are detailed in
Table~\ref{tab_5}.
\begin{table} \centering \caption{The Number of
images in each category (Air.means airport, Rail. railway, Comme.
commercial, Inter. intersection, Temp. template, Univ.
university)}\begin{tabular}{|l|c|c|c|c|c|c|} \hline
Categroy &Air.   &Comme.  &Industrial   &Inter.   &Park   \\
\hline Number &306   &262   &206   &302   &129    \\\hline
Categroy  &Rail.     &Seaport   &Soccer   &Temp.   &Univ. \\
\hline Number &115   &126   &128   &218    &305      \\
\hline\end{tabular} \label{tab_5}
\end{table}

\section{Conclusions}
Aerial image categorization is an important component in
{artificial intelligence and remote
sensing~\cite{add4,add5}.} In this paper, a new geometric
discriminative feature {is proposed for}
aerial image recognition. Both the local features and their
geometric property are taken into account to
{describe an aerial image.} A region
connected graph (RCG) is defined to encode the geometric property
and the color intensity of an aerial image. Then, the frequent
structures are mined statistically from the training RCGs. The
refined structures are further selected from the frequent
structures toward being highly discriminative and low redundant.
{Given a new aerial image, its geometric
discriminative features are extracted guided by the refined
structures,} They are further quantized into a vector for
SVM~\cite{ksvm} classification. We evaluated the effectiveness of
our approach on both the public and our own data sets.

\section{Appendix}
Ideally, we want a perfect segmentation algorithm with two merits:
First, each segmented region represents a semantic
object/component. Second, the segmentation algorithm is
parameter-free. Thus, we can apply it to segment thousands of
training images once for all, without human-interactive parameter
tuning. Unfortunately, for the first merit, the high-level
features in those semantics-exploited segmentation methods are
usually designed manually and data set dependent, which is not
consistent with the fully-automated and data set independent
framework of the proposed method; besides, to learn semantics,
semantics-exploited segmentation methods typically require
well-annotated training images, however, the large number of
training aerial images used in our experiment are online crawled
and human annotation is laborious. For the second merit, those
semantic-exploited segmentation methods are usually complicated
and there are several important user-controlled parameters.
Therefore, we can only use those data-driven segmentation methods,
where no semantics are explored and typically contain one tuning
parameter. Those well-known data-driven segmentation algorithms
can be divided into two groups. The first group algorithms need
the number of segmented regions as input, such as k-means and
normalized cut; however, there is no uniform segmented region
number on different images because different images usually
contain different number of components. The second group
algorithms require some tolerance bound as input, such as the
similarity tolerance between spatially neighboring segmented
regions. Compared with segmented region number, we empirically
found that the tolerance bound is more flexible to tune.
Therefore, in our approach, we chose the second group data-driven
segmentation methods. After some experimental comparison, we found
that the unsupervised fuzzy clustering~\footnote{Matlab codes:
https://mywebspace.wisc.edu/pwang6/personal/} outperforms several
tolerance bound-based segmentation algorithms, such as graph-based
segmentation~\footnote{C++ codes:
http://www.cs.brown.edu/~pff/segment/}. Thus, we choose
unsupervised fuzzy clustering in our approach.

\end{document}